\documentclass[11pt,a4paper]{article}
\usepackage[hyperref]{acl2017}
\usepackage{times}
\usepackage{latexsym}
\usepackage{graphicx}
\usepackage{amsmath}
\usepackage{amsfonts}
\usepackage[linesnumbered]{algorithm2e}
\usepackage{url}
\usepackage{tabularx}
\usepackage{multirow}

\newcommand{\argmax}{\arg\!\max}

\aclfinalcopy % Uncomment this line for the final submission
\title{AMR-to-text Generation with Synchronous Node Replacement Grammar}

\author{Linfeng Song, Xiaochang Peng, Yue Zhang, Zhiguo Wang \and Daniel Gildea \\
  Department of Computer Science, University of Rochester, Rochester, NY 14627 \\
  IBM T.J. Watson Research Center, Yorktown Heights, NY 10598 \\
  Singapore University of Technology and Design}

\date{}

\begin{document}
\maketitle
\begin{abstract}
This paper addresses the task of AMR-to-text generation by leveraging synchronous node replacement grammar.
During training, graph-to-string rules are learned using a heuristic extraction algorithm.
At test time, a graph transducer is applied to collapse input AMRs and generate output sentences.
Evaluated on a standard benchmark, our method gives the state-of-the-art result.
\end{abstract}

\section{Introduction}

Abstract Meaning Representation (AMR) \cite{banarescu-EtAl:2013:LAW7-ID} is a semantic formalism encoding the meaning of a sentence as a rooted, directed graph.
AMR uses a graph to represent meaning, where nodes (such as ``boy'', ``want-01'') represent concepts, and edges (such as ``ARG0'', ``ARG1'') represent relations between concepts.
Encoding many semantic phenomena into a graph structure, AMR is useful for NLP tasks such as machine translation \cite{jones2012semantics,tamchyna-quirk-galley:2015:S2MT}, question answering \cite{mitra2015addressing}, summarization \cite{takase-EtAl:2016:EMNLP2016} and event detection \cite{li-EtAl:2015:CNewsStory}.

AMR-to-text generation is challenging as function words and syntactic structures are abstracted away, making an AMR graph correspond to multiple realizations.
Despite much literature so far on text-to-AMR parsing \cite{flanigan2014discriminative,wang-xue-pradhan:2015:NAACL-HLT,peng2015synchronous,vanderwende2015amr,pust2015parsing,artzi-lee-zettlemoyer:2015:EMNLP,groschwitz-koller-teichmann:2015:ACL-IJCNLP,goodman-vlachos-naradowsky:2016:P16-1,zhou-EtAl:2016:EMNLP20163,peng-EtAl:2017:EACLlong1},
there has been little work on AMR-to-text generation \cite{jeff2016amrgen,song-EtAl:2016:EMNLP2016,pourdamghani-knight-hermjakob:2016:INLG}.

\newcite{jeff2016amrgen} transform a given AMR graph into a spanning tree, before translating it to a sentence using a tree-to-string transducer.
Their method leverages existing machine translation techniques, capturing hierarchical correspondences between the spanning tree and the surface string.
However, it suffers from error propagation since the output is constrained given a spanning tree due to the projective correspondence between them.
Information loss in the graph-to-tree transformation step cannot be recovered.
\newcite{song-EtAl:2016:EMNLP2016} directly generate sentences using graph-fragment-to-string rules.
They cast the task of finding a sequence of disjoint rules to transduce an AMR graph into a sentence as a traveling salesman problem, using local features and a language model to rank candidate sentences.
However, their method does not learn hierarchical structural correspondences between AMR graphs and strings.

\begin{figure}
\centering
\includegraphics[scale=.6]{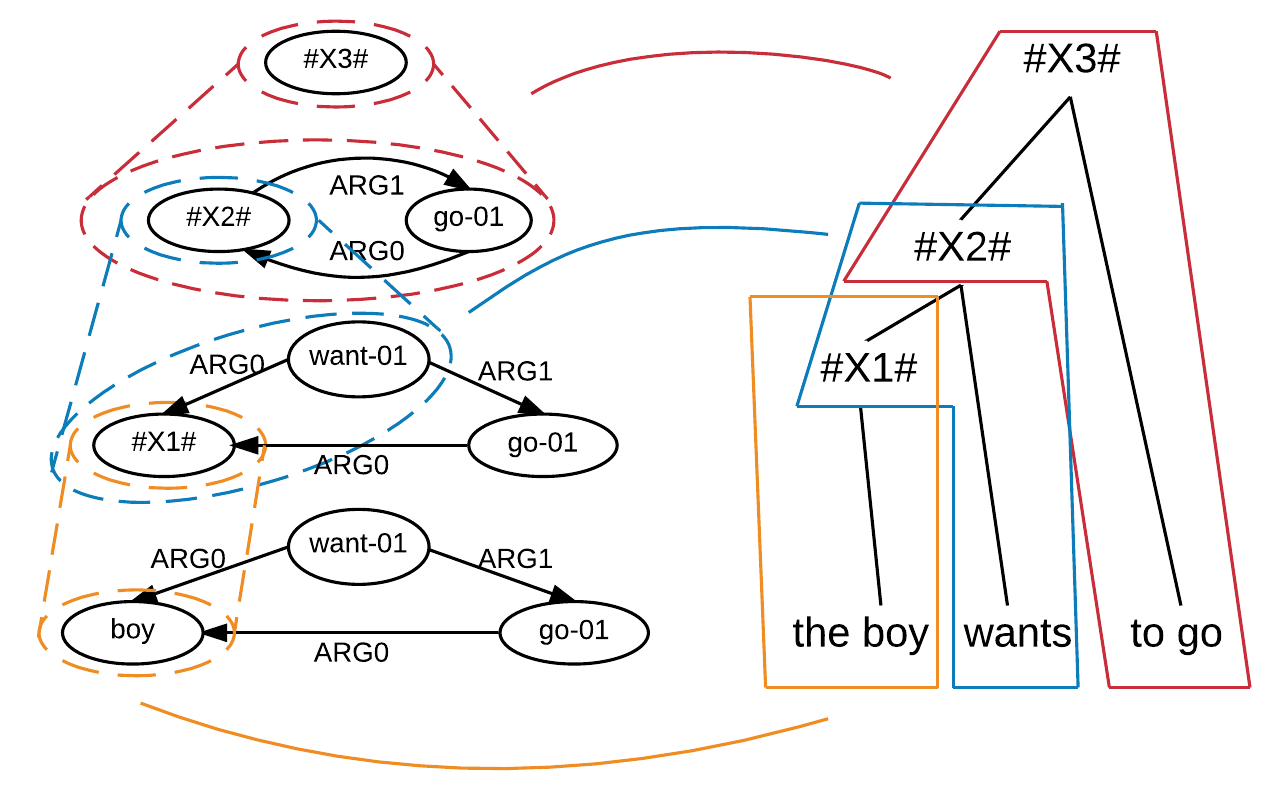}
\caption{Graph-to-string derivation.}
\label{fig:derivation}
\end{figure}

\begin{figure*}[t]
\centering
\includegraphics[scale=0.65]{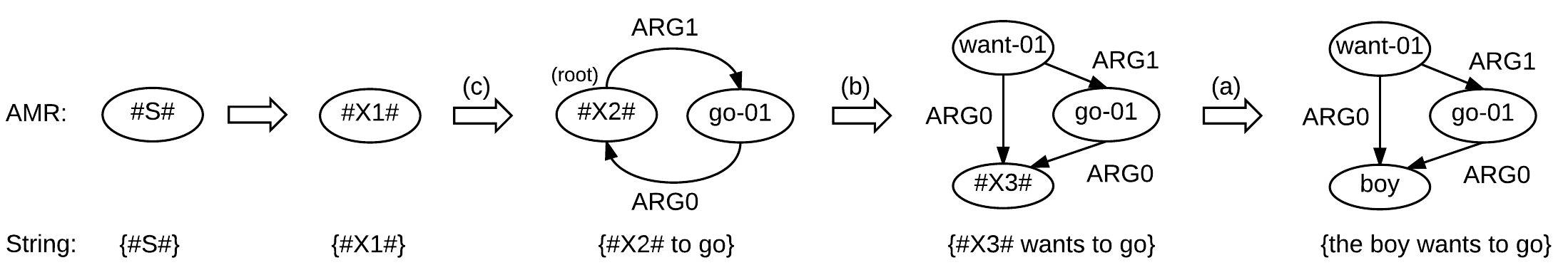}
\caption{Example deduction procedure}
\label{fig:deduction_proc}
\end{figure*}

We propose to leverage the advantages of hierarchical rules without suffering from graph-to-tree errors by directly learning graph-to-string rules.
As shown in Figure \ref{fig:derivation}, we learn a synchronous node replacement grammar (NRG) from a corpus of aligned AMR and sentence pairs.
At test time, we apply a graph transducer to collapse input AMR graphs and generate output strings according to the learned grammar.
Our system makes use of a log-linear model with real-valued features, tuned using MERT \cite{och:2003:ACL}, and beam search decoding.
It gives a BLEU score of 25.62 on LDC2015E86, which is the state-of-the-art on this dataset.

\begin{table}[t] \small
\centering
 \begin{tabular}{|lll|} 
 \hline
  ID. & $F$ & $E$ \\
 \hline
  (a)                 & (b / boy)               & the boy \\
 \hline
 \multirow{2}{*}{(b)} & (w / want-01            & \multirow{3}{*}{\#X\# wants} \\
                      & ~~~~~:ARG0 (X / \#X\#)) & \\
 \hline
 \multirow{3}{*}{(c)} & (X / \#X\#              & \multirow{3}{*}{\#X\# to go} \\
                      & ~~~~~:ARG1 (g / go-01   & \\
                      & ~~~~~~~~~:ARG0 X))      & \\
 \hline
 \multirow{2}{*}{(d)} & (w / want-01            & \multirow{3}{*}{the boy wants} \\
                      & ~~~~~:ARG0 (b / boy))   & \\
 \hline
 \end{tabular}
 \caption{Example rule set} \label{tab:ruleset}
\end{table}

\section{Synchronous Node Replacement Grammar}

\subsection{Grammar Definition}

A synchronous node replacement grammar (NRG) is a rewriting formalism: $G=\langle N, \Sigma, \Delta, P, S \rangle$, where $N$ is a finite set of nonterminals, $\Sigma$ and $\Delta$ are finite sets of terminal symbols for the source and target sides, respectively.
$S \in N$ is the start symbol, and $P$ is a finite set of productions.
Each instance of $P$ takes the form $X_i \rightarrow (\langle F, E\rangle,\sim)$, where $X_i \in N$ is a nonterminal node, $F$ is a rooted, connected AMR fragment with edge labels over $\Sigma$ and node labels over $N \cup \Sigma$, $E$ is a corresponding target string over $N \cup \Delta$ and $\sim$ denotes the alignment of nonterminal symbols between $F$ and $E$.
A classic NRG \cite[Chapter~1]{Engelfriet:1997} also defines $C$, which is an embedding mechanism defining how $F$ is connected to the rest of the graph when replacing $X_i$ with $F$ on the graph. 
Here we omit defining $C$ and allow arbitrary connections.\footnote{This may over generate, but does not affect our case, as in our bottom-up decoding procedure (section \ref{sec:model}) when $F$ is replaced with $X_i$, nodes previously connected to $F$ are re-connected to $X_i$}
Following \newcite{chiang:2005:ACL}, we use only one nonterminal $X$ in addition to $S$, and use subscripts to distinguish different non-terminal instances.

Figure \ref{fig:deduction_proc} shows an example derivation process for the sentence ``the boy wants to go'' given the rule set in Table \ref{tab:ruleset}. 
Given the start symbol $S$, which is first replaced with $X_1$, rule (c) is applied to generate ``$X_2$ to go'' and its AMR counterpart.
Then rule (b) is used to generate ``$X_3$ wants'' and its AMR counterpart from $X_2$.
Finally, rule (a) is used to generate ``the boy'' and its AMR counterpart from $X_3$.
Our graph-to-string rules are inspired by synchronous grammars for machine translation \cite{wu1997stochastic,yamada-knight:2002:ACL,Gildea-acl03,chiang:2005:ACL,huang2006statistical,liu-liu-lin:2006:COLACL,shen-xu-weischedel:2008:ACLMain,xie-mi-liu:2011:EMNLP,meng-EtAl:2013:EMNLP}. 

\subsection{Induced Rules}

\begin{algorithm}[t] \small
 \KwData{training corpus $C$}
 \KwResult{rule instances $R$}
 $R$ $\leftarrow$ []\;
 \For{$(Sent,AMR,\sim)$ \textbf{in} $C$}{
   $R_{cur}$ $\leftarrow$ \textsc{FragmentExtract}($Sent$,$AMR$,$\sim$)\;
   \For{$r_i$ \textbf{in} $R_{cur}$}{
     $R$.\textsc{append}($r_i$) \;
   	 \For{$r_j$ \textbf{in} $R_{cur}/\{r_i\}$}{
        \If{$r_i$.\textsc{Contains}$(r_j)$}{
           $r_{ij}$ $\leftarrow$ $r_i$.\textsc{collapse}($r_j$)\;
           $R$.\textsc{append}($r_{ij}$) \;
        }
     }
   }
 }
 \caption{Rule extraction}
 \label{algo:dump}
\end{algorithm}

There are three types of rules in our system, namely \emph{induced rules}, \emph{concept rules} and \emph{graph glue rules}.
Here we first introduce induced rules, which are obtained by a two-step procedure on a training corpus.
Shown in Algorithm \ref{algo:dump}, the first step is to extract a set of initial rules from training $\langle$sentence, AMR, $\sim$$\rangle$\footnote{$\sim$ denotes alignment between words and AMR labels.} pairs (Line 2) using the phrase-to-graph-fragment extraction algorithm of \newcite{peng2015synchronous} (Line 3). 
Here an \emph{initial rule} contains only terminal symbols in both $F$ and $E$.
As a next step, we match between pairs of initial rules $r_i$ and $r_j$, and generate $r_{ij}$ by collapsing $r_i$ with $r_j$, if $r_i$ contains $r_j$ (Line 6-8).
Here $r_i$ contains $r_j$, if $r_j.F$ is a subgraph of $r_i.F$ and $r_j.E$ is a sub-phrase of $r_i.E$.
When collapsing $r_i$ with $r_j$, we replace the corresponding subgraph in $r_i.F$ with a new non-terminal node, and the sub-phrase in $r_i.E$ with the same non-terminal.
For example, we obtain rule (b) by collapsing (d) with (a) in Table \ref{tab:ruleset}.
All initial and generated rules are stored in a rule list $R$ (Lines 5 and 9), which will be further normalized to obtain the final induced rule set.

\subsection{Concept Rules and Glue Rules}

In addition to induced rules, we adopt concept rules \cite{song-EtAl:2016:EMNLP2016} and graph glue rules to ensure existence of derivations.
For a concept rule, $F$ is a single node in the input AMR graph, and $E$ is a morphological string of the node concept.
A concept rule is used in case no induced rule can cover the node.
We refer to the verbalization list\footnote{http://amr.isi.edu/download/lists/verbalization-list-v1.06.txt}
and AMR guidelines\footnote{https://github.com/amrisi/amr-guidelines} for creating more complex concept rules.
For example, one concept rule created from the verbalization list is ``(k / keep-01 :ARG1 (p / peace)) $|||$ peacekeeping''.

Inspired by \newcite{chiang:2005:ACL}, we define graph glue rules to concatenate non-terminal nodes connected with an edge, when no induced rules can be applied. 
Three glue rules are defined for each type of edge label.
Taking the edge label ``ARG0'' as an example, we create the following glue rules:
\vspace{1mm}

\begin{small}
\noindent 
\begin{tabular}{lll} 
\hline
  ID. & $F$ & $E$ \\
\hline
  $r_1$ & (X1 / \#X1\# :ARG0 (X2 / \#X2\#)) & \#X1\#~~\#X2\# \\
  $r_2$ & (X1 / \#X1\# :ARG0 (X2 / \#X2\#)) & \#X2\#~~\#X1\# \\
  $r_3$ & (X1 / \#X1\# :ARG0 X1)            & \#X1\# \\
\hline
\end{tabular}
\end{small}

\vspace{1mm}
\noindent
where for both $r_1$ and $r_2$, $F$ contains two non-terminal nodes with a directed edge connecting them, and $E$ is the concatenation the two non-terminals in either the monotonic or the inverse order.
For $r_3$, $F$ contains one non-terminal node with a self-pointing edge, and $E$ is the non-terminal.
With concept rules and glue rules in our final rule set, it is easily guaranteed that there are legal derivations for any input AMR graph. 

\section{Model}
\label{sec:model}

We adopt a log-linear model for scoring search hypotheses. Given an input AMR graph, we find the highest scored derivation $t^{\ast}$ from all possible derivations $t$:
\begin{equation}
t^{\ast} = \argmax_{t} \exp\sum_i w_i f_i(g,t)\textrm{,}
\label{eq:log_linear}
\end{equation}
where $g$ denotes the input AMR, $f_i(\cdot,\cdot)$ and $w_i$ represent a feature and the corresponding weight, respectively.
The feature set that we adopt includes phrase-to-graph and graph-to-phrase translation probabilities and their corresponding lexicalized translation probabilities (section \ref{sec:trans_prob}), language model score, word count, rule count, reordering model score (section \ref{sec:reorder}) and moving distance (section \ref{sec:moving_dis}).
The language model score, word count and phrase count features are adopted from SMT \cite{koehn2003statistical,chiang:2005:ACL}.

We perform bottom-up search to transduce input AMRs to surface strings.
Each hypothesis contains the current AMR graph, translations of collapsed subgraphs, the feature vector and the current model score.
Beam search is adopted, where hypotheses with the same number of collapsed edges and nodes are put into the same beam.

\subsection{Translation Probabilities}
\label{sec:trans_prob}

Production rules serve as a basis for scoring hypotheses.
We associate each synchronous NRG rule $n \rightarrow (\langle F, E \rangle,\sim)$ with a set of probabilities.
First, phrase-to-fragment translation probabilities are defined based on maximum likelihood estimation (MLE), as shown in Equation \ref{eq:phr}, where $c_{\langle F, E \rangle}$ is the fractional count of $\langle F, E \rangle$.
\begin{equation}
p(F|E)=\frac{c_{\langle F,E \rangle}}{\sum_{F'}c_{\langle F',E \rangle}}
\label{eq:phr}
\end{equation}
In addition, lexicalized translation probabilities are defined as:
\begin{equation}
p_w(F|E)=\prod_{l \in F}{\sum_{w \in E} p(l|w)}
\label{eq:lex}
\end{equation}
Here $l$ is a label (including both edge labels such as ``ARG0'' and concept labels such as ``want-01'') in the AMR fragment $F$, and $w$ is a word in the phrase $E$.
Equation \ref{eq:lex} can be regarded as a ``soft'' version of the lexicalized translation probabilities adopted by SMT, which picks the alignment yielding the maximum lexicalized probability for each translation rule.
%\begin{equation}
%p_w(f|e)=\max_a\prod_{f_i \in f}{p(f_i|a_{f_i})}
%\label{eq:smt}
%\end{equation}
In addition to $p(F|E)$ and $p_w(F|E)$, we use features in the reverse direction, namely $p(E|F)$ and $p_w(E|F)$, the definitions of which are omitted as they are consistent with Equations \ref{eq:phr} and \ref{eq:lex}, respectively.
The probabilities associated with concept rules and glue rules are manually set to 0.0001.

\subsection{Reordering Model}
\label{sec:reorder}

Although the word order is defined for induced rules, it is not the case for glue rules.
We learn a reordering model that helps to decide whether the translations of the nodes should be monotonic or inverse given the directed connecting edge label.
The probabilistic model using smoothed counts is defined as:
\begin{multline}
p(M|h,l,t)=\\
\frac{1.0+\sum_{h}\sum_{t}c(h,l,t,M)}{2.0+\sum_{o\in\{M,I\}}\sum_{h}\sum_{t}c(h,l,t,o)}
\label{eq:reorder}
\end{multline}
$c(h,l,t,M)$ is the count of monotonic translations of head $h$ and tail $t$, connected by edge $l$.

\subsection{Moving Distance}
\label{sec:moving_dis}

The moving distance feature captures the distances between the subgraph roots of two consecutive rule matches in the decoding process, which controls a bias towards collapsing nearby subgraphs consecutively.

\section{Experiments}

\subsection{Setup}

We use LDC2015E86 as our experimental dataset, which contains 16833 training, 1368 dev and 1371 test instances. 
Each instance contains a sentence, an AMR graph and the alignment generated by a heuristic aligner.
Rules are extracted from the training data, and model parameters are tuned on the dev set. 
For tuning and testing, we filter out sentences with more than 30 words, resulting in 1103 dev instances and 1055 test instances. 
We train a 4-gram language model (LM) on gigaword (LDC2011T07), and use BLEU \cite{papineni2002bleu} as the evaluation metric.
MERT is used \cite{och:2003:ACL} to tune model parameters on $k$-best outputs on the devset, where $k$ is set 50.
%\footnote{We also tried MIRA \cite{chiang-marton-resnik:2008:EMNLP} and PRO \cite{hopkins-may:2011:EMNLP}, neither of which further improves the performance.}

We investigate the effectiveness of rules and features by ablation tests: 
``NoInducedRule'' does not adopt induced rules, 
``NoConceptRule'' does not adopt concept rules, 
``NoMovingDistance'' does not adopt the moving distance feature, and 
``NoReorderModel'' disables the reordering model.
Given an AMR graph, if \emph{NoConceptRule} cannot produce a legal derivation, we concatenate existing translation fragments into a final translation, 
and if a subgraph can not be translated, the empty string is used as the output.
We also compare our method with previous works, in particular JAMR-gen \cite{jeff2016amrgen} and TSP-gen \cite{song-EtAl:2016:EMNLP2016}, on the same dataset.%\footnote{The BLEU of TSP-gen is from the original paper \cite{song-EtAl:2016:EMNLP2016}, and that of JAMR-gen is produced by the authors (Flanigan et al.,) on our dataset}

\begin{table}[t] \small
\centering
 \begin{tabular}{|l|l|l|} 
 \hline
 System & Dev & Test \\
 \hline
 \hline
 TSP-gen  & 21.12 & 22.44 \\
 \hline
 JAMR-gen & 23.00 & 23.00 \\
 \hline
 All              & \textbf{25.24} & \textbf{25.62} \\
 NoInducedRule    & 16.75 & 17.43 \\
 NoConceptRule    & 23.99 & 24.86 \\
 NoMovingDistance & 23.48 & 24.06 \\
 NoReorderModel   & 25.09 & 25.43 \\
 \hline
 \end{tabular}
 \caption{Main results.} \label{tab:rst}
\end{table}

\subsection{Main results}

The results are shown in Table \ref{tab:rst}.
First, \emph{All} outperforms all baselines.
\emph{NoInducedRule} leads to the greatest performance drop compared with \emph{All}, demonstrating that induced rules play a very important role in our system.
On the other hand, \emph{NoConceptRule} does not lead to much performance drop.
This observation is consistent with the observation of \newcite{song-EtAl:2016:EMNLP2016} for their TSP-based system.
\emph{NoMovingDistance} leads to a significant performance drop, empirically verifying the fact that the translations of nearby subgraphs are also close.
Finally, \emph{NoReorderingModel} does not affect the performance significantly, which can be because the most important reordering patterns are already covered by the hierarchical induced rules.
Compared with \emph{TSP-gen} and \emph{JAMR-gen}, our final model \emph{All} improves the BLEU from 22.44 and 23.00 to 25.62, showing the advantage of our model.
To our knowledge, this is the best result reported so far on the task.

\subsection{Grammar analysis}

We have shown the effectiveness of our synchronous node replacement grammar (SNRG) on the AMR-to-text generation task.
Here we further analyze our grammar as it is relatively less studied than the hyperedge replacement grammar (HRG) \cite{drewes1997hyperedge}.

\begin{figure}
\centering
\includegraphics[scale=.34]{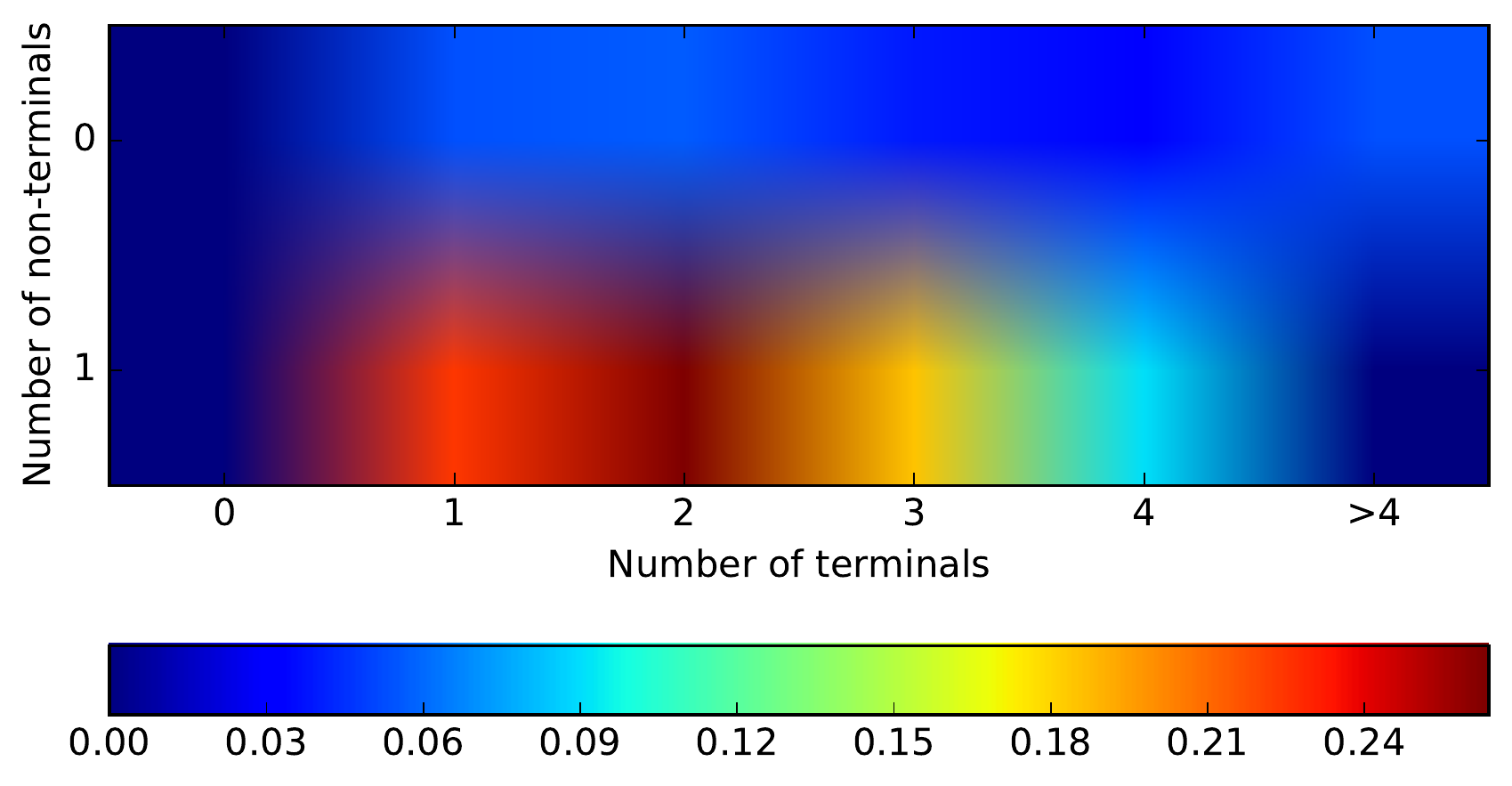}
\caption{Statistics on the right-hand side.}
\label{fig:stat}
\end{figure}

\vspace{1mm}
\noindent
\textbf{Statistics on the whole rule set}
%\subsubsection{Statistics on the whole rule set}

\vspace{1mm}
\noindent
We first categorize our rule set by the number of terminals and nonterminals in the AMR fragment $F$, and show the percentages of each type in Figure \ref{fig:stat}.
Each rule contains at most 1 nonterminal, as we collapse each initial rule only once.
First of all, the percentage of rules containing nonterminals are much more than those without nonterminals,
as we collapse each pair of initial rules (in Algorithm \ref{algo:dump}) and the results can be quadratic the number of initial rules.
In addition, most rules are small containing 1 to 3 terminals, meaning that they represent small pieces of meaning and are easier to matched on a new AMR graph.
Finally, there are a few large rules, which represent complex meaning.

\begin{table} \small
\centering
\begin{tabular}{lccc}
\hline
            & Glue   & Nonterminal & Terminal \\
\hline
1-best      & 30.0\% & 30.1\% & 39.9\% \\
\hline
\end{tabular}
\caption{Rules used for decoding.}
\label{tab:rule_used}
\end{table}

\vspace{1mm}
\noindent
\textbf{Statistics on the rules used for decoding}
%\subsubsection{Statistics on the rules used for decoding}

\vspace{1mm}
\noindent
In addition, we collect the rules that our well-tuned system used for generating the 1-best output on the testset, and categorize them into 3 types: (1) glue rules, (2) nonterminal rules, which are not glue rules but contain nonterminals on the right-hand side and (3) terminal rules, whose right-hand side only contain terminals.
Over the rules used on the 1-best result, more than 30\% are non-terminal rules, showing that the induced rules play an important role.
On the other hand, 30\% are glue rules.
The reason is that the data sparsity for graph grammars is more severe than string-based grammars (such as CFG), as the graph structures are more complex than strings.
Finally, terminal rules take the largest percentage, while most are \emph{induced rules}, but not \emph{concept rules}.

\vspace{1mm}
\noindent
\textbf{Rule examples}
%\subsubsection{Rule examples}

\vspace{1mm}
\noindent
Finally, we show some rules in Table \ref{tab:example}, where $F$ and $E$ are the right-hand-side AMR fragment and phrase, respectively.
For the first rule, the root of $F$ is a verb (``give-01'') whose subject is a nonterminal and object is a AMR fragment ``(p / person :ARG0-of (u / use-01))'', which means ``user''.
So it is easy to see that the corresponding phrase $E$ conveys the same meaning.
For the second rule, ``(s3 / stay-01 :accompanier (i / i))'' means ``stay with me'', which is also covered by its phrase.

\begin{table} \small
\centering
\begin{tabular}{ll}
\hline
\multirow{4}{*}{$F$:}& (g / give-01 \\
                     & ~~~~~~~~:ARG0 (X1 / \#X1\#) \\
                     & ~~~~~~~~:ARG2 (p / person \\
                     & ~~~~~~~~~~~~~~~~:ARG0-of (u / use-01))) \\
$E$:                 & \#X1\# has given users an \\
\hline
%\multirow{4}{*}{$F$:}& (c / consider-01 \\
%                     & ~~~~~:ARG0 (p / person \\
%                     & ~~~~~~~~~~:mod (s / some)) \\
%                     & ~~~~~:ARG1 (X1 / \#X1\#)) \\
%$E$:                 & some people consider it a \#X1\# \\
%\hline
\multirow{3}{*}{$F$:}& (X1 / \#X1\# \\
                     & ~~~~~~~~:ARG2 (s3 / stay-01 :ARG1 X1 \\
                     & ~~~~~~~~~~~~~~~~:accompanier (i / i))) \\
$E$:                 & \#X1\# staying with me \\
\hline
%\multirow{3}{*}{$F$:}& (a / attract-01 \\
%                     & ~~~~~:ARG1 (X1 / \#X1\#) \\
%                     & ~~~~~:manner (p / physical)) \\
%$E$:                 & \#X1\# wont be attracted to them physically \\
%\hline
%\multirow{3}{*}{$F$:}& (X1 / \#X1\# \\
%                     & ~~~~~:ARG1 (g / generate-01 \\
%                     & ~~~~~~~~~~:ARG1 (m / much))) \\
%$E$:                 & \#X1\# generate much of \\
%\hline
\end{tabular}
\caption{Example rules.}
\label{tab:example}
\end{table}

\begin{table} \small
\centering
\begin{tabularx}{0.48\textwidth}{X}
\hline
(u / understand-01 \\
~~~~:ARG0 (y / you) \\
~~~~:ARG1 (t2 / thing \\
~~~~~~~~:ARG1-of (f2 / feel-01 \\
~~~~~~~~~~~~:ARG0 (p2 / person \\
~~~~~~~~~~~~~~~~:example (p / person :wiki - \\
~~~~~~~~~~~~~~~~~~~~:name (t / name :op1 ``TMT'') \\
~~~~~~~~~~~~~~~~~~~~:location (c / city :wiki ``Fairfax,\_Virginia'' \\
~~~~~~~~~~~~~~~~~~~~~~~~:name (f / name :op1 ``Fairfax'')))))) \\
~~~~:time (n / now)) \\
\hline
\textbf{Trans:} now, you have to understand that people feel about such as tmt fairfax \\
\hline
\textbf{Ref:} now you understand how people like tmt in fairfax feel .\\
\hline
\end{tabularx}
\caption{Generation example.}
\label{tab:example_gen}
\end{table}

\subsection{Generation example}

Finally, we show an example in Table \ref{tab:example_gen}, where the top is the input AMR graph, and the bottom is the generation result.
Generally, most of the meaning of the input AMR are correctly translated, such as ``:example'', which means ``such as'', and ``thing'', which is an abstract concept and should not be translated,
while there are a few errors, such as ``that'' in the result should be ``what'', and there should be an ``in'' between ``tmt'' and ``fairfax''.

\section{Conclusion}

We showed that synchronous node replacement grammar is useful for AMR-to-text generation by developing a system that learns a synchronous NRG in the training time, and applies a graph transducer to collapse input AMR graphs and generate output strings according to the learned grammar at test time.
Our method performs better than the previous systems, empirically proving the advantages of our graph-to-string rules.

\section*{Acknowledgement}

This work was funded by a Google Faculty Research Award. 
Yue Zhang is funded by NSFC61572245 and T2MOE201301 from Singapore Ministry of Education.

%\bibliography{acl2017}
\bibliography{acl2017}

\begin{thebibliography}{}
\expandafter\ifx\csname natexlab\endcsname\relax\def\natexlab#1{#1}\fi

\bibitem[{Artzi et~al.(2015)Artzi, Lee, and
  Zettlemoyer}]{artzi-lee-zettlemoyer:2015:EMNLP}
Yoav Artzi, Kenton Lee, and Luke Zettlemoyer. 2015.
\newblock Broad-coverage {CCG} semantic parsing with {AMR}.
\newblock In {\em Conference on Empirical Methods in Natural Language
  Processing (EMNLP-15)\/}. pages 1699--1710.

\bibitem[{Banarescu et~al.(2013)Banarescu, Bonial, Cai, Georgescu, Griffitt,
  Hermjakob, Knight, Koehn, Palmer, and
  Schneider}]{banarescu-EtAl:2013:LAW7-ID}
Laura Banarescu, Claire Bonial, Shu Cai, Madalina Georgescu, Kira Griffitt, Ulf
  Hermjakob, Kevin Knight, Philipp Koehn, Martha Palmer, and Nathan Schneider.
  2013.
\newblock Abstract meaning representation for sembanking.
\newblock In {\em Proceedings of the 7th Linguistic Annotation Workshop and
  Interoperability with Discourse\/}. pages 178--186.

\bibitem[{Chiang(2005)}]{chiang:2005:ACL}
David Chiang. 2005.
\newblock A hierarchical phrase-based model for statistical machine
  translation.
\newblock In {\em Proceedings of the 43rd Annual Meeting of the Association for
  Computational Linguistics (ACL-05)\/}. Ann Arbor, Michigan, pages 263--270.

\bibitem[{Drewes et~al.(1997)Drewes, Kreowski, and Habel}]{drewes1997hyperedge}
Frank Drewes, Hans-J{\"o}rg Kreowski, and Annegret Habel. 1997.
\newblock Hyperedge replacement, graph grammars.
\newblock {\em Handbook of Graph Grammars\/} 1:95--162.

\bibitem[{Engelfriet and Rozenberg(1997)}]{Engelfriet:1997}
J.~Engelfriet and G.~Rozenberg. 1997.
\newblock Node replacement graph grammars.
\newblock In Grzegorz Rozenberg, editor, {\em Handbook of Graph Grammars and
  Computing by Graph Transformation\/}, World Scientific Publishing Co., Inc.,
  River Edge, NJ, USA, pages 1--94.

\bibitem[{Flanigan et~al.(2016)Flanigan, Dyer, Smith, and
  Carbonell}]{jeff2016amrgen}
Jeffrey Flanigan, Chris Dyer, Noah~A. Smith, and Jaime Carbonell. 2016.
\newblock Generation from abstract meaning representation using tree
  transducers.
\newblock In {\em Proceedings of the 2016 Meeting of the North American chapter
  of the Association for Computational Linguistics (NAACL-16)\/}. pages
  731--739.

\bibitem[{Flanigan et~al.(2014)Flanigan, Thomson, Carbonell, Dyer, and
  Smith}]{flanigan2014discriminative}
Jeffrey Flanigan, Sam Thomson, Jaime Carbonell, Chris Dyer, and Noah~A. Smith.
  2014.
\newblock A discriminative graph-based parser for the abstract meaning
  representation.
\newblock In {\em Proceedings of the 52nd Annual Meeting of the Association for
  Computational Linguistics (ACL-14)\/}. pages 1426--1436.

\bibitem[{Gildea(2003)}]{Gildea-acl03}
Daniel Gildea. 2003.
\newblock Loosely tree-based alignment for machine translation.
\newblock In {\em Proceedings of the 41th Annual Conference of the Association
  for Computational Linguistics (ACL-03)\/}. Sapporo, Japan, pages 80--87.

\bibitem[{Goodman et~al.(2016)Goodman, Vlachos, and
  Naradowsky}]{goodman-vlachos-naradowsky:2016:P16-1}
James Goodman, Andreas Vlachos, and Jason Naradowsky. 2016.
\newblock Noise reduction and targeted exploration in imitation learning for
  abstract meaning representation parsing.
\newblock In {\em Proceedings of the 54th Annual Meeting of the Association for
  Computational Linguistics (ACL-16)\/}. Berlin, Germany, pages 1--11.

\bibitem[{Groschwitz et~al.(2015)Groschwitz, Koller, and
  Teichmann}]{groschwitz-koller-teichmann:2015:ACL-IJCNLP}
Jonas Groschwitz, Alexander Koller, and Christoph Teichmann. 2015.
\newblock Graph parsing with s-graph grammars.
\newblock In {\em Proceedings of the 53rd Annual Meeting of the Association for
  Computational Linguistics (ACL-15)\/}. Beijing, China, pages 1481--1490.

\bibitem[{Huang et~al.(2006)Huang, Knight, and Joshi}]{huang2006statistical}
Liang Huang, Kevin Knight, and Aravind Joshi. 2006.
\newblock Statistical syntax-directed translation with extended domain of
  locality.
\newblock In {\em Proceedings of Association for Machine Translation in the
  Americas (AMTA-2006)\/}. pages 66--73.

\bibitem[{Jones et~al.(2012)Jones, Andreas, Bauer, Hermann, and
  Knight}]{jones2012semantics}
Bevan Jones, Jacob Andreas, Daniel Bauer, Karl~Moritz Hermann, and Kevin
  Knight. 2012.
\newblock Semantics-based machine translation with hyperedge replacement
  grammars.
\newblock In {\em Proceedings of the International Conference on Computational
  Linguistics (COLING-12)\/}. pages 1359--1376.

\bibitem[{Koehn et~al.(2003)Koehn, Och, and Marcu}]{koehn2003statistical}
Philipp Koehn, Franz~Josef Och, and Daniel Marcu. 2003.
\newblock Statistical phrase-based translation.
\newblock In {\em Proceedings of the 2003 Meeting of the North American chapter
  of the Association for Computational Linguistics (NAACL-03)\/}. pages 48--54.

\bibitem[{Li et~al.(2015)Li, Nguyen, Cao, and
  Grishman}]{li-EtAl:2015:CNewsStory}
Xiang Li, Thien~Huu Nguyen, Kai Cao, and Ralph Grishman. 2015.
\newblock Improving event detection with abstract meaning representation.
\newblock In {\em Proceedings of the First Workshop on Computing News
  Storylines\/}. Beijing, China, pages 11--15.

\bibitem[{Liu et~al.(2006)Liu, Liu, and Lin}]{liu-liu-lin:2006:COLACL}
Yang Liu, Qun Liu, and Shouxun Lin. 2006.
\newblock Tree-to-string alignment template for statistical machine
  translation.
\newblock In {\em Proceedings of the 44th Annual Meeting of the Association for
  Computational Linguistics (ACL-06)\/}. Sydney, Australia, pages 609--616.

\bibitem[{Meng et~al.(2013)Meng, Xie, Song, L\"{u}, and
  Liu}]{meng-EtAl:2013:EMNLP}
Fandong Meng, Jun Xie, Linfeng Song, Yajuan L\"{u}, and Qun Liu. 2013.
\newblock Translation with source constituency and dependency trees.
\newblock In {\em Conference on Empirical Methods in Natural Language
  Processing (EMNLP-13)\/}. Seattle, Washington, USA, pages 1066--1076.

\bibitem[{Mitra and Baral(2015)}]{mitra2015addressing}
Arindam Mitra and Chitta Baral. 2015.
\newblock Addressing a question answering challenge by combining statistical
  methods with inductive rule learning and reasoning.
\newblock In {\em Proceedings of the National Conference on Artificial
  Intelligence (AAAI-16)\/}.

\bibitem[{Och(2003)}]{och:2003:ACL}
Franz~Josef Och. 2003.
\newblock Minimum error rate training in statistical machine translation.
\newblock In {\em Proceedings of the 41st Annual Meeting of the Association for
  Computational Linguistics (ACL-03)\/}. Sapporo, Japan, pages 160--167.

\bibitem[{Papineni et~al.(2002)Papineni, Roukos, Ward, and
  Zhu}]{papineni2002bleu}
Kishore Papineni, Salim Roukos, Todd Ward, and Wei-Jing Zhu. 2002.
\newblock {BLEU}: a method for automatic evaluation of machine translation.
\newblock In {\em Proceedings of the 40th Annual Meeting of the Association for
  Computational Linguistics (ACL-02)\/}. pages 311--318.

\bibitem[{Peng et~al.(2015)Peng, Song, and Gildea}]{peng2015synchronous}
Xiaochang Peng, Linfeng Song, and Daniel Gildea. 2015.
\newblock A synchronous hyperedge replacement grammar based approach for {AMR}
  parsing.
\newblock In {\em Proceedings of the Nineteenth Conference on Computational
  Natural Language Learning (CoNLL-15)\/}. pages 731--739.

\bibitem[{Peng et~al.(2017)Peng, Wang, Gildea, and
  Xue}]{peng-EtAl:2017:EACLlong1}
Xiaochang Peng, Chuan Wang, Daniel Gildea, and Nianwen Xue. 2017.
\newblock Addressing the data sparsity issue in neural amr parsing.
\newblock In {\em Proceedings of the 15th Conference of the European Chapter of
  the Association for Computational Linguistics (EACL-17)\/}. Valencia, Spain,
  pages 366--375.

\bibitem[{Pourdamghani et~al.(2016)Pourdamghani, Knight, and
  Hermjakob}]{pourdamghani-knight-hermjakob:2016:INLG}
Nima Pourdamghani, Kevin Knight, and Ulf Hermjakob. 2016.
\newblock Generating {English} from abstract meaning representations.
\newblock In {\em International Conference on Natural Language Generation
  (INLG-16)\/}. Edinburgh, UK, pages 21--25.

\bibitem[{Pust et~al.(2015)Pust, Hermjakob, Knight, Marcu, and
  May}]{pust2015parsing}
Michael Pust, Ulf Hermjakob, Kevin Knight, Daniel Marcu, and Jonathan May.
  2015.
\newblock Parsing {English} into abstract meaning representation using
  syntax-based machine translation.
\newblock In {\em Conference on Empirical Methods in Natural Language
  Processing (EMNLP-15)\/}. pages 1143--1154.

\bibitem[{Shen et~al.(2008)Shen, Xu, and
  Weischedel}]{shen-xu-weischedel:2008:ACLMain}
Libin Shen, Jinxi Xu, and Ralph Weischedel. 2008.
\newblock A new string-to-dependency machine translation algorithm with a
  target dependency language model.
\newblock In {\em Proceedings of the 46th Annual Meeting of the Association for
  Computational Linguistics (ACL-08)\/}. Columbus, Ohio, pages 577--585.

\bibitem[{Song et~al.(2016)Song, Zhang, Peng, Wang, and
  Gildea}]{song-EtAl:2016:EMNLP2016}
Linfeng Song, Yue Zhang, Xiaochang Peng, Zhiguo Wang, and Daniel Gildea. 2016.
\newblock {AMR}-to-text generation as a traveling salesman problem.
\newblock In {\em Conference on Empirical Methods in Natural Language
  Processing (EMNLP-16)\/}. Austin, Texas, pages 2084--2089.

\bibitem[{Takase et~al.(2016)Takase, Suzuki, Okazaki, Hirao, and
  Nagata}]{takase-EtAl:2016:EMNLP2016}
Sho Takase, Jun Suzuki, Naoaki Okazaki, Tsutomu Hirao, and Masaaki Nagata.
  2016.
\newblock Neural headline generation on abstract meaning representation.
\newblock In {\em Conference on Empirical Methods in Natural Language
  Processing (EMNLP-16)\/}. Austin, Texas, pages 1054--1059.

\bibitem[{Tamchyna et~al.(2015)Tamchyna, Quirk, and
  Galley}]{tamchyna-quirk-galley:2015:S2MT}
Ale\v{s} Tamchyna, Chris Quirk, and Michel Galley. 2015.
\newblock A discriminative model for semantics-to-string translation.
\newblock In {\em Proceedings of the 1st Workshop on Semantics-Driven
  Statistical Machine Translation (S2MT 2015)\/}. Beijing, China, pages 30--36.

\bibitem[{Vanderwende et~al.(2015)Vanderwende, Menezes, and
  Quirk}]{vanderwende2015amr}
Lucy Vanderwende, Arul Menezes, and Chris Quirk. 2015.
\newblock An {AMR} parser for {English}, {French}, {German}, {Spanish} and
  {Japanese} and a new {AMR}-annotated corpus.
\newblock In {\em Proceedings of the 2015 Meeting of the North American chapter
  of the Association for Computational Linguistics (NAACL-15)\/}. pages 26--30.

\bibitem[{Wang et~al.(2015)Wang, Xue, and
  Pradhan}]{wang-xue-pradhan:2015:NAACL-HLT}
Chuan Wang, Nianwen Xue, and Sameer Pradhan. 2015.
\newblock A transition-based algorithm for {AMR} parsing.
\newblock In {\em Proceedings of the 2015 Meeting of the North American chapter
  of the Association for Computational Linguistics (NAACL-15)\/}. pages
  366--375.

\bibitem[{Wu(1997)}]{wu1997stochastic}
Dekai Wu. 1997.
\newblock Stochastic inversion transduction grammars and bilingual parsing of
  parallel corpora.
\newblock {\em Computational linguistics\/} 23(3):377--403.

\bibitem[{Xie et~al.(2011)Xie, Mi, and Liu}]{xie-mi-liu:2011:EMNLP}
Jun Xie, Haitao Mi, and Qun Liu. 2011.
\newblock A novel dependency-to-string model for statistical machine
  translation.
\newblock In {\em Conference on Empirical Methods in Natural Language
  Processing (EMNLP-11)\/}. Edinburgh, Scotland, UK., pages 216--226.

\bibitem[{Yamada and Knight(2002)}]{yamada-knight:2002:ACL}
Kenji Yamada and Kevin Knight. 2002.
\newblock A decoder for syntax-based statistical {MT}.
\newblock In {\em Proceedings of the 40th Annual Meeting of the Association for
  Computational Linguistics (ACL-02)\/}. Philadelphia, Pennsylvania, USA, pages
  303--310.

\bibitem[{Zhou et~al.(2016)Zhou, Xu, Uszkoreit, QU, Li, and
  Gu}]{zhou-EtAl:2016:EMNLP20163}
Junsheng Zhou, Feiyu Xu, Hans Uszkoreit, Weiguang QU, Ran Li, and Yanhui Gu.
  2016.
\newblock {AMR} parsing with an incremental joint model.
\newblock In {\em Conference on Empirical Methods in Natural Language
  Processing (EMNLP-16)\/}. Austin, Texas, pages 680--689.

\end{thebibliography}
\bibliographystyle{acl_natbib}

\end{document}